\pdfoutput=1
\documentclass{article}

\usepackage{microtype}
\usepackage{graphicx}
\usepackage{booktabs} 

\usepackage{hyperref}

\usepackage{color}
\usepackage{subcaption}
\usepackage{amsmath}
\usepackage{amssymb}
\usepackage{listings}
\usepackage{appendix}

\definecolor{deepcarrotorange}{rgb}{0.91, 0.41, 0.17}
\newcommand{\ignore}[1]{}

\iftrue 
\newcommand{\fixme}[1]{{\textcolor{red}{[~FIXME:~#1~]}}}
\newcommand{\note}[1]{{\bf [~NOTE:~#1~]}}
\newcommand{\todo}[1]{{\bf [~TODO:~#1~]}}
\newcommand{\jichuan}[1]{{\textcolor{red}{[~JICHUAN:~#1~]}}}
\newcommand{\grant}[1]{{\textcolor{blue}{[~GRANT:~#1~]}}}
\newcommand{\heiner}[1]{{\textcolor{blue}{[~HEINER:~#1~]}}}
\newcommand{\milad}[1]{{\textcolor{red}{[~MILAD:~#1~]}}}
\newcommand{\christos}[1]{{\textcolor{deepcarrotorange}{[~CHRISTOS:~#1~]}}}
\newcommand{\partha}[1]{{\textcolor{red}{[~PARTHA:~#1~]}}}
\newcommand{\kevin}[1]{{\textcolor{red}{[~KEVIN:~#1~]}}}
\newcommand{\tmp}[1]{{\textcolor{red}{#1~}}}
\else
\newcommand{\fixme}[1]{}
\newcommand{\note}[1]{}
\newcommand{\todo}[1]{}
\newcommand{\jichuan}[1]{}
\newcommand{\grant}[1]{}
\newcommand{\heiner}[1]{}
\newcommand{\milad}[1]{}
\newcommand{\christos}[1]{}
\newcommand{\partha}[1]{}
\newcommand{\kevin}[1]{}
\newcommand{\tmp}[1]{}
\fi

\newcommand{\bx}{\mathbf{x}}

\newcommand{\bh}{\mathbf{h}}
\newcommand{\bc}{\mathbf{c}}
\newcommand{\bb}{\mathbf{b}}
\newcommand{\bi}{\mathbf{i}}
\newcommand{\bforget}{\mathbf{f}}
\newcommand{\bo}{\mathbf{o}}

\usepackage[accepted]{icml2018}

\icmltitlerunning{Learning Memory Access Patterns}

\begin{document}

\twocolumn[
\icmltitle{Learning Memory Access Patterns}



\icmlsetsymbol{equal}{*}

\begin{icmlauthorlist}
\icmlauthor{Milad Hashemi}{google}
\icmlauthor{Kevin Swersky}{google}
\icmlauthor{Jamie A. Smith}{google}
\icmlauthor{Grant Ayers}{stanford,equal}
\icmlauthor{Heiner Litz}{ucsc,equal}
\icmlauthor{Jichuan Chang}{google}
\icmlauthor{Christos Kozyrakis}{stanford}
\icmlauthor{Parthasarathy Ranganathan}{google}
\end{icmlauthorlist}

\icmlaffiliation{google}{Google}
\icmlaffiliation{stanford}{Stanford University}
\icmlaffiliation{ucsc}{University of California, Santa Cruz}

\icmlcorrespondingauthor{Milad Hashemi}{miladh@google.com}
\icmlcorrespondingauthor{Kevin Swersky}{kswersky@google.com}

\icmlkeywords{Machine Learning, Computer Architecture, Cache Prefetching, LSTM, Neural Networks}

\vskip 0.3in
]



\printAffiliationsAndNotice{\icmlEqualContribution} 

\begin{abstract}
The explosion in workload complexity and the recent slow-down in Moore's law scaling call for new approaches towards efficient computing. Researchers are now beginning to use recent advances in machine learning in software optimizations, augmenting or replacing traditional heuristics and data structures. However, the space of machine learning for computer hardware architecture is only lightly explored. In this paper, we demonstrate the potential of deep learning to address the von Neumann bottleneck of memory performance. We focus on the critical problem of learning memory access patterns, with the goal of constructing accurate and efficient memory prefetchers. We relate contemporary prefetching strategies to n-gram models in natural language processing, and show how recurrent neural networks can serve as a drop-in replacement. On a suite of challenging benchmark datasets, we find that neural networks consistently demonstrate superior performance in terms of precision and recall. This work represents the first step towards practical neural-network based prefetching, and opens a wide range of exciting directions for machine learning in computer architecture research.
\end{abstract}

\section{Introduction}
\label{sec:introduction}

The proliferation of machine learning, and more recently deep learning, in real-world applications has been made possible by an exponential increase in compute capabilities, largely driven by advancements in hardware design. To maximize the effectiveness of a given design, computer architecture often involves the use of prediction and heuristics. Prefetching is a canonical example of this, where instructions or data are brought into much faster storage well in advance of their required usage. 

Prefetching addresses a critical bottleneck in von Neumann computers: computation is orders of magnitude faster than accessing memory. This problem is known as the memory wall \cite{wul:mck95}, and modern applications can spend over 50\% of all compute cycles waiting for data to arrive from memory \cite{kozyrakis2010,Ferdman:2012,Kanev:2015}. To mitigate the memory wall, microprocessors use a hierarchical memory system, with small and fast memory close to the processor (i.e.,~caches), and large yet slower memory farther away. Prefetchers predict when to fetch what data into cache to reduce memory latency, and the key towards effective prefetching is to attack the difficult problem of predicting memory access patterns.

Predictive optimization such as prefetching is one form of speculation. Modern microprocessors leverage numerous types of predictive structures to issue speculative requests with the aim of increasing performance. Historically, most predictors in hardware are table-based.  That is, future events are expected to correlate with past history tracked in lookup tables (implemented as memory arrays). These memory arrays are sized based on the working set, or amount of information that the application actively uses. However, the working sets of modern datacenter workloads are orders of magnitude larger than those of traditional workloads such as \textit{SPEC CPU2006} and continue to grow~\cite{Ayers:2018, Ferdman:2012, gutierrez:2011, bursty:2016}. This trend poses a significant challenge, as prediction accuracy drops sharply when the working set is larger than the predictive table. Scaling predictive tables with fast-growing working sets is difficult and costly for hardware implementation.

Neural networks have emerged as a powerful technique to address sequence prediction problems, such as those found in natural language processing (NLP) and text understanding~\cite{bengio2003neural,mikolov2010recurrent,mikolov2013word2vec}. Simple perceptrons have even been deployed in hardware (e.g., SPARC T4 processor~\cite{golla:jordan:2011}), to handle branch prediction \cite{jim:lin01}. Yet, exploring the effectiveness of sequential learning algorithms in microarchitectural designs is still an open area of research.

In this paper, we explore the utility of sequence-based neural networks in microarchitectural systems. In particular, given the challenge of the memory wall, we apply sequence learning to the difficult problem of prefetching. 

Prefetching is fundamentally a regression problem. The output space, however, is both vast and extremely sparse, making it a poor fit for standard regression models. We take inspiration from recent works in image and audio generation that discretize the space, namely PixelRNN and Wavenet \cite{vanoord2016pixelrnn, oord2016wavenet}. Discretization makes prefetching more analogous to neural language models, and we leverage it as a starting point for building neural prefetchers. We find that we can successfully model the output space to a degree of accuracy that makes neural prefetching a very distinct possibility. On a number of benchmark datasets, we find that recurrent neural networks significantly outperform the state-of-the-art of traditional hardware prefetchers. We also find that our results are interpretable. Given a memory access trace, we show that the RNN is able to discern semantic information about the underlying application.

\section{Background}
\label{sec:background}

\subsection{Microarchitectural Data Prefetchers}

Prefetchers are hardware structures that predict future memory accesses from past history. They can largely be separated into two categories: stride prefetchers and correlation prefetchers. Stride prefetchers are commonly implemented in modern processors and lock onto stable, repeatable deltas (differences between subsequent memory addresses)  \cite{gin77, jou90, pal:kes94}. For example, given an access pattern that adds four to a memory address every time (0, 4, 8, 12), a stride prefetcher will learn that delta and try to prefetch ahead of the demand stream, launching parallel accesses to potential future address targets (16, 20, 24) up to a set prefetch distance.

Correlation prefetchers try to learn patterns that may repeat, but are not as consistent as a single stable delta \cite{cha:ree95,lai:fid01,somogyi:2006,rot:mos98}. They store the past history of memory accesses in large tables and are better at predicting more irregular patterns than stride prefetchers. Examples of correlation prefetchers include Markov prefetchers \cite{jos:gru97}, GHB prefetchers \cite{nes:smith04}, and more recent work that utilizes larger in-memory structures \cite{jain2013linearizing}. Correlation prefetchers require large, costly tables, and are typically not implemented in modern multi-core processors.

\subsection{Recurrent Neural Networks}
Deep learning has become the model-class of choice for many sequential prediction problems. Notably, speech recognition \cite{hinton2012deep} and natural language processing \cite{mikolov2010recurrent}. In particular, RNNs are a preferred choice for their ability to model long-range dependencies. LSTMs \cite{hochreiter1997long} have emerged as a popular RNN variant that deals with training issues in standard RNNs, by propagating the internal state additively instead of multiplicatively.
An LSTM is composed of a hidden state $\bh$ and a cell state $\bc$, along with input $\bi$, forget $\bforget$, and output gates $\bo$ that dictate what information gets stored and propagated to the next timestep. At timestep $N$, input $\bx_N$ is presented to the LSTM, and the LSTM states are computed using the following process:

\begin{enumerate}
	\item Compute the input, forget, and output gates
	\begin{align*}
		\bi_N &= \sigma(W_i[\bx_N, \bh_{N-1}] + \bb_i) \\
		\bforget_N &= \sigma(W_f[\bx_N, \bh_{N-1}] + \bb_f) \\
		\bo_N &= \sigma(W_o[\bx_N, \bh_{N-1}] + \bb_o)
	\end{align*}
	\item Update the cell state
	\begin{align*}
		\bc_N &= \bforget_N \odot \bc_{N-1} + \bi_N \odot \mathrm{tanh}(W_c[\bx_N, \bh_{N-1}] + \bb_c)
	\end{align*}
	\item Compute the LSTM hidden (output) state
	\begin{align*}
		\bh_N &= \bo_N \odot \mathrm{tanh}(\bc_N)
	\end{align*}
\end{enumerate}

Where $[\bx_N, \bh_{N-1}]$ represents the concatenation of the current input and previous hidden state, $\odot$ represents element-wise multiplication, and $\sigma(u) = \frac{1}{1 + \exp(-u)}$ is the sigmoid non-linearity.

The above process forms a single LSTM layer, where $W_{\{i,f,o,c\}}$ are the weights of the layer, and $\bb_{\{i,f,o,c\}}$ are the biases. LSTM layers can be further stacked so that the output of one LSTM layer at time $N$ becomes the input to another LSTM layer at time $N$. It is analogous to having multiple layers in a feed-forward neural network, and allows greater modeling flexibility with relatively few extra parameters.

\section{Problem Formulation}
\label{sec:problem_formulation}
\subsection{Prefetching as a Prediction Problem}
\label{sec:pref_prediction}
Prefetching is the process of predicting future memory accesses that will miss in the on-chip cache and access memory based on past history. Each of these memory addresses are generated by a memory instruction (a load/store). Memory instructions are a subset of all instructions that interact with the addressable memory of the computer system. 

Many hardware proposals use two features to make these prefetching decisions: the sequence of caches miss addresses that have been observed so far and the sequence of instruction addresses, also known as program counters (PCs), that are associated with the instruction that generated each of the cache miss addresses. 

PCs are unique tags, that is each PC is unique to a particular instruction that has been compiled from a particular function in a particular code file. PC sequences can inform the model of patterns in the control flow of higher level code, while the miss address sequence informs the model of which address to prefetch next. In modern computer systems, both of these features are represented as 64-bit integers.

Therefore, an initial model could use two input features at a given timestep $N$. It could use the address and PC that generated a cache miss at that timestep to predict the address of the miss at timestep $N+1$.

\begin{figure}[ht]
	\centering
	\includegraphics[width=\columnwidth]{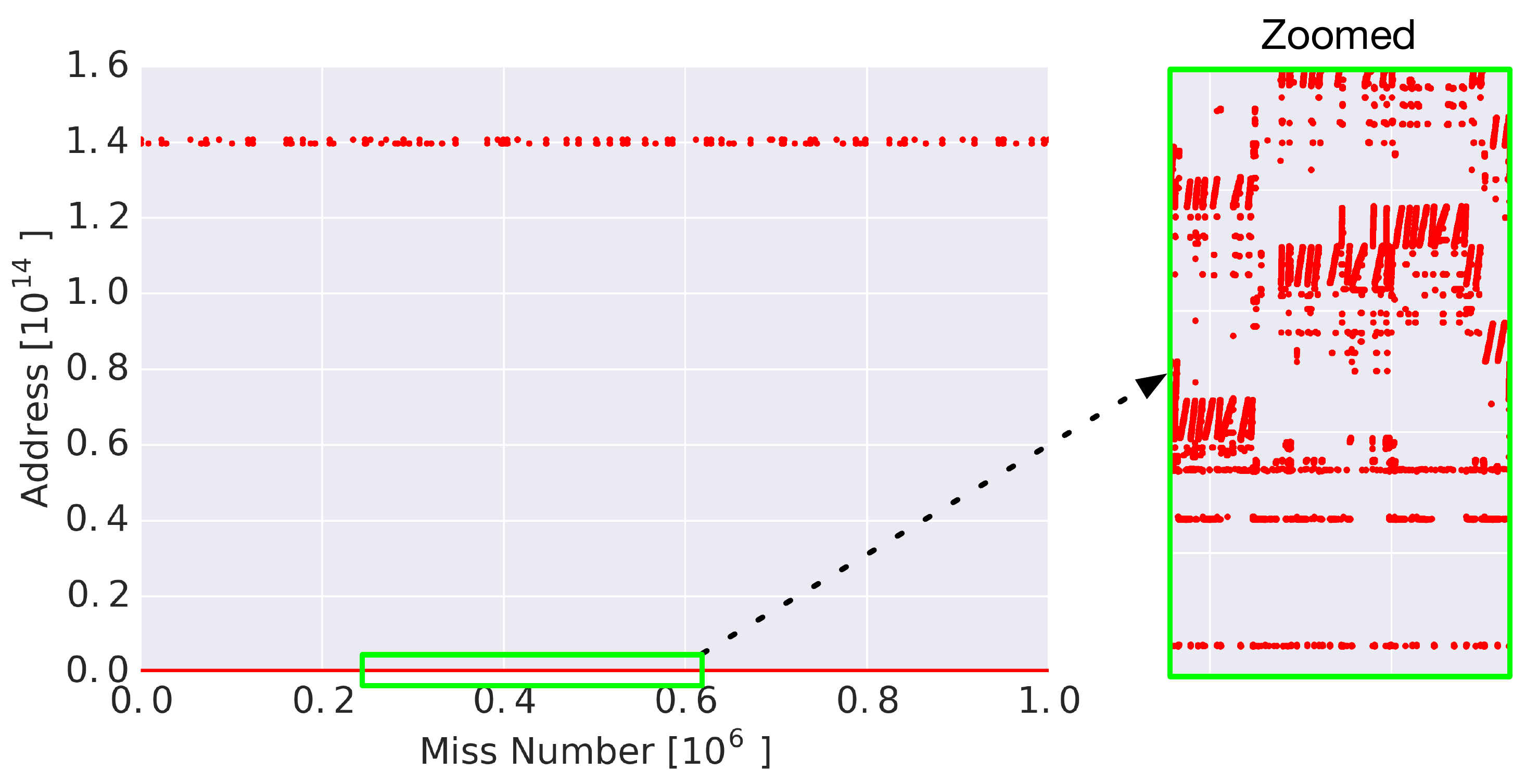}
	\caption{Cache miss addresses on the \textit{omnetpp} dataset, demonstrating sparse access patterns at multiple scales.}
	\label{fig:omnetpp_zoom}
\end{figure}

However, one concern quickly becomes apparent: the address space of an application is extremely sparse. In our training data with O(100M) cache misses, only O(10M) unique cache block miss addresses appear on average out of the entire $\mathrm{2}^\mathrm{64}$ physical address space. This is further displayed when we plot an example trace from \textit{omnetpp} (a benchmark from the standard \textit{SPEC CPU2006} benchmark suite \cite{spec}) in Figure \ref{fig:omnetpp_zoom}, where the red datapoints are cache miss addresses\footnote{Cache miss addresses are from a three level simulated cache hierarchy with a 32 KB L1, 256 KB L2, and 1.25 MB Last Level Cache (LLC), similar to a single thread context from an Intel Broadwell microprocessor.}. The wide range and severely multi-modal nature of this space makes it a challenge for time-series regression models. For example, neural networks tend work best with normalized inputs, however when normalizing this data, the finite precision floating-point representation results in a significant loss of information. This issue affects modeling at both the input and output levels, and we will describe several approaches to deal with both aspects.

\subsection{Prefetching as Classification}
\label{sec:prefetching_as_classification}

Rather than treating the prefetching problem as regression, we opt to treat the address space as a large, discrete vocabulary, and perform classification. This is analogous to next-word or character prediction in natural language processing. The extreme sparsity of the space, and the fact that some addresses are much more commonly accessed than others, means that the effective vocabulary size can actually be manageable for RNN models. Additionally, the model gains flexibility by being able to produce multi-modal outputs, compared to unimodal regression techniques that assume e.g., a Gaussian likelihood. This idea of treating the output prediction problem as one of classification instead of regression has been successfully used in image \cite{vanoord2016pixelrnn} and audio generation \cite{oord2016wavenet}.

However, there are $\mathrm{2}^{64}$ possible softmax targets, so a quantization scheme is necessary. Importantly, in order to be useful, a prefetch must be within a cache line to be completely accurate, usually within 64 bytes. There is a second order benefit if it is within a page, usually 4096 bytes, but even predicting at the page level would leave $\mathrm{2}^{52}$ possible targets. In \cite{oord2016wavenet}, they predict 16-bit integer values from an acoustic signal. To avoid having to apply a softmax over $\mathrm{2}^\mathrm{16}$ values, they apply a non-linear quantization scheme to reduce the space to 256 categories. This form of quantization is inappropriate for our purposes, as it decreases the resolution of addresses towards the extremes of the address space, whereas in prefetching we need high resolution in every area where addresses are used.

Luckily, programs tend to behave in predictable ways, so only a relatively small (but still large in absolute numbers), and consistent set of addresses are ever seen. Our primary quantization scheme is to therefore create a vocabulary of common addresses during training, and to use this as the set of targets during testing. This reduces the coverage, as there may be addresses at test time that are not seen during training time, however we will show that for reasonably-sized vocabularies, we capture a significant proportion of the space. The second approach we explore is to cluster the addresses using clustering on the address space. This is akin to an adaptive form of non-linear quantization.

Due to dynamic side-effects such as address space layout randomization (ASLR), different runs of the same program will lead to different raw address accesses \cite{team2003pax}. However, given a layout, the program will behave in a consistent manner. Therefore, one potential strategy is to predict deltas, $\Delta_N = \mathrm{Addr}_{N+1} - \mathrm{Addr}_{N}$, instead of addresses directly. These will remain consistent across program executions, and come with the benefit that the number of uniquely occurring deltas is often orders of magnitude smaller than uniquely occurring addresses. This is shown in Table \ref{table:dataset_stats}, where we show the number of unique PCs, addresses, and deltas across a suite of program trace datasets. We also show the number of unique addresses and deltas required to achieve 50\% coverage. In almost all cases, this is much smaller when considering deltas. In our models, we therefore use deltas as inputs instead of raw addresses.

\begin{table*}[ht!]
	\begin{center}
	\caption{Program trace dataset statistics. M stands for million.}
	\label{table:dataset_stats}
		\begin{tabular}{ |c|r|r|r|r|r|r| } 
			\hline
			Dataset & \# Misses & \# PC & \# Addrs & \# Deltas & \# Addrs 50\% mass & \# Deltas 50\% mass \\
			\hline
			gems				&	500M & 3278		& 13.11M	& 2.47M   & 4.28M & 18 \\
			astar				&	500M & 211		& 0.53M	  & 1.77M   & 0.06M & 15 \\
			bwaves			&	491M & 893    & 14.20M	& 3.67M   & 3.03M & 2 \\
			lbm 			 	&	500M & 55     & 6.60M		& 709     & 3.06M & 9 \\
			leslie3d		&	500M & 2554   & 1.23M   & 0.03M   & 0.23M & 15 \\
			libquantum	&	470M & 46     & 0.52M		& 30      & 0.26M & 1 \\
			mcf					&	500M & 174		& 27.41M  & 30.82M  & 0.07M & 0.09M \\
			milc 				&	500M & 898		& 3.74M		& 9.68M   & 0.87M & 46 \\
			omnetpp			&	449M & 976		& 0.71M   & 5.01M   & 0.12M & 4613 \\
			soplex 			&	500M & 1218		& 3.49M   & 5.27M   & 1.04M & 10 \\
			sphinx			&	283M & 693		& 0.21M		& 0.37M   & 0.03M & 3 \\
			websearch	&	500M & 54600	& 77.76M	& 96.41M  & 0.33M & 5186 \\
			\hline
		\end{tabular}
	\end{center}
	\vspace{-.2in}
\end{table*}

\section{Models}
In this section we introduce two LSTM-based prefetching models. The first version is analogous to a standard language model, while the second exploits the structure of the memory access space in order to reduce the vocabulary size and reduce the model memory footprint.

\subsection{Embedding LSTM}

Suppose we restricted the output vocabulary size in order to only model the most frequently occurring deltas. According to Table \ref{table:dataset_stats}, the size of the vocabulary required in order to obtain at best 50\% accuracy is usually O(1000) or less, well within the capabilities of standard language models. Our first model therefore restricts the output vocabulary size to a large, but feasible 50,000 of the most frequent, unique deltas. For the input vocabulary, we include all deltas as long as they appear in the dataset at least 10 times. Expanding the vocabulary beyond this is challenging, both computationally and statistically. We leave an exploration of approaches like the hierarchical softmax \cite{mnih2009scalable} to future work.

\begin{figure}[ht]
    \centering
	\includegraphics[width=\columnwidth]{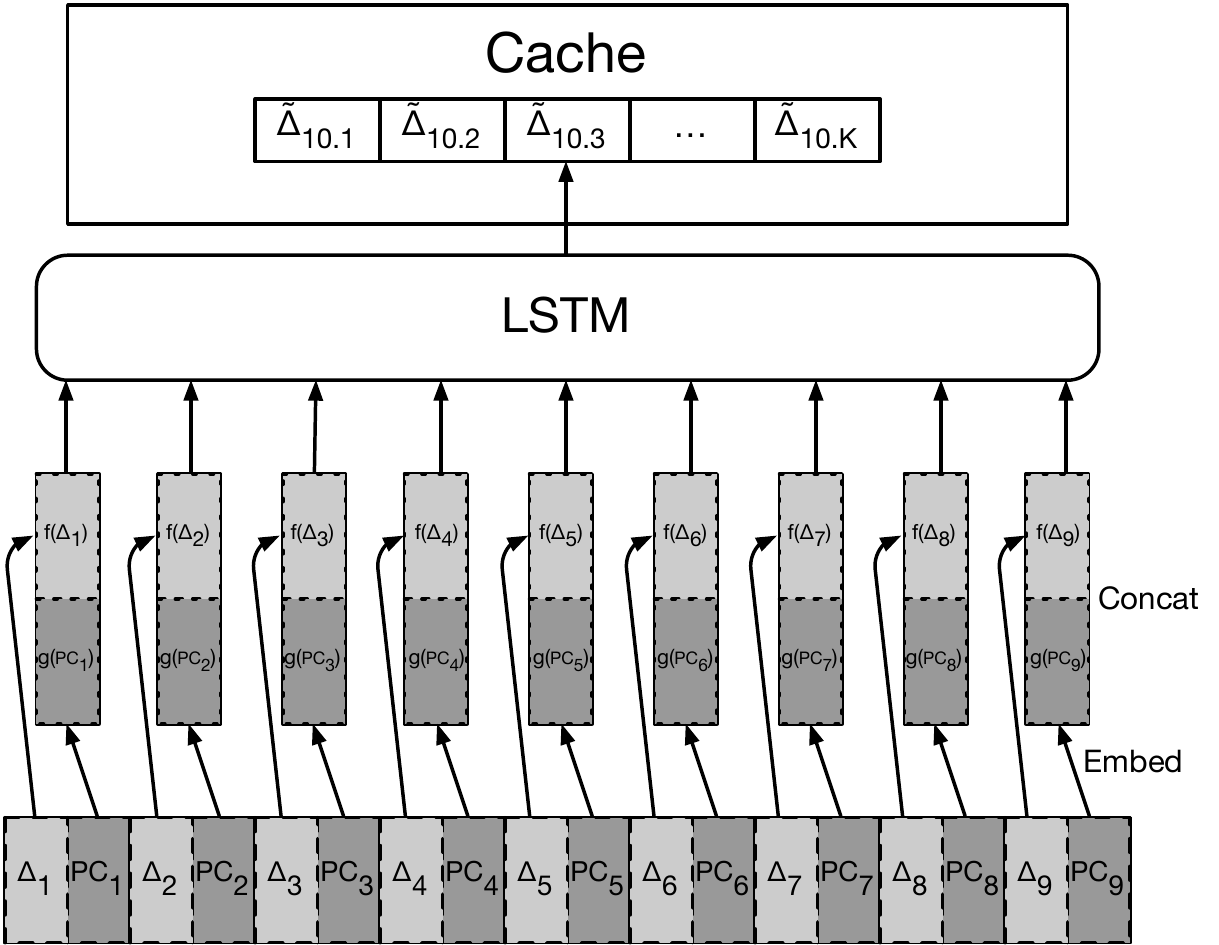}
	\caption{The embedding LSTM model. $f$ and $g$ represent embedding functions.}
	\label{fig:embedding_lstm}
\end{figure}

We refer to this model as the embedding LSTM, as illustrated in Figure~\ref{fig:embedding_lstm}. It uses a categorical (one-hot) representation for both the input and output deltas. At timestep $N$, the input $\mathrm{PC}_N$ and $\Delta_N$ are individually embedded and then the embeddings are concatenated and fed as inputs to a two-layer LSTM. The LSTM then performs classification over the delta vocabulary, and the $K$ highest-probability deltas are chosen for prefetching \footnote{Directly predicting probabilites is another advantage that classification provides over traditional hardware.}.

In a practical implementation, a prefetcher can return several predictions. This creates a trade-off, where more predictions increases the probability of a cache hit at the next timestep, but potentially removes other useful items from the cache~\footnote{This trade-off is subtle, because some of the addresses may end up being useful at future timesteps.}. We opt to prefetch the top-10 predictions of the LSTM at each timestep. Other possibilities that we do not explore here include using a beam-search to predict the next $n$ deltas, or to learn to directly predict $N$ to $N+n$ steps ahead in one forward pass of the LSTM.

There are several limitations to this approach. First, a large vocabulary increases the model's computational and storage footprint. Second, truncating the vocabulary necessarily puts a ceiling on the accuracy of the model. Finally, dealing with rarely occurring deltas is non-trivial, as they will be seen relatively few times during training. This is known in NLP as the rare word problem \cite{luong2015rare}.

\subsection{Clustering + LSTM}

We hypothesize that much of the interesting interaction between addresses occurs locally in address space. As one example, data structures like structs and arrays tend to be stored in contiguous blocks, and accessed repeatedly. In this model, we exploit this idea to design a prefetcher that very carefully models \emph{local} context, whereas the embedding LSTM models both local and global context.

By looking at narrower regions of the address space, we can see that there is indeed rich local context. We took the set of addresses from \textit{omnetpp} and clustered them into 6 different regions using k-means. We show two of the clusters in Figure \ref{fig:kmeans}, and the rest can be found in the appendix.

\begin{figure}
    \centering
    \includegraphics[width=\columnwidth]{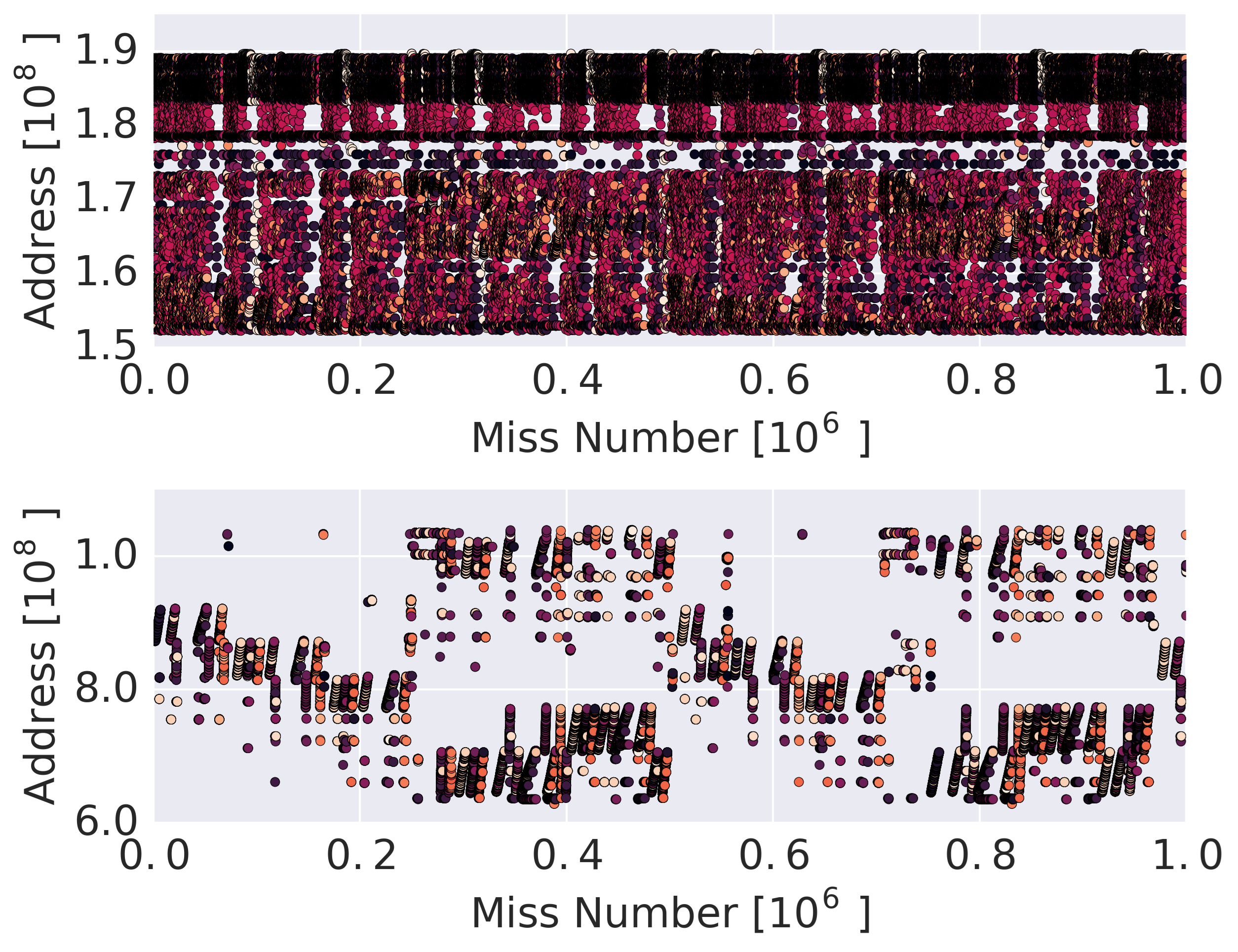}
    
    \caption{Two of six k-means clusters on the \textit{omnetpp} benchmark dataset. Memory accesses are colored according to the PC that generated them.}\label{fig:omnetpp_clusters}
    \label{fig:kmeans}
\end{figure}

\begin{figure*}[t!]
	\centering
	\begin{subfigure}{0.5\textwidth}
    	\includegraphics[width=\textwidth]{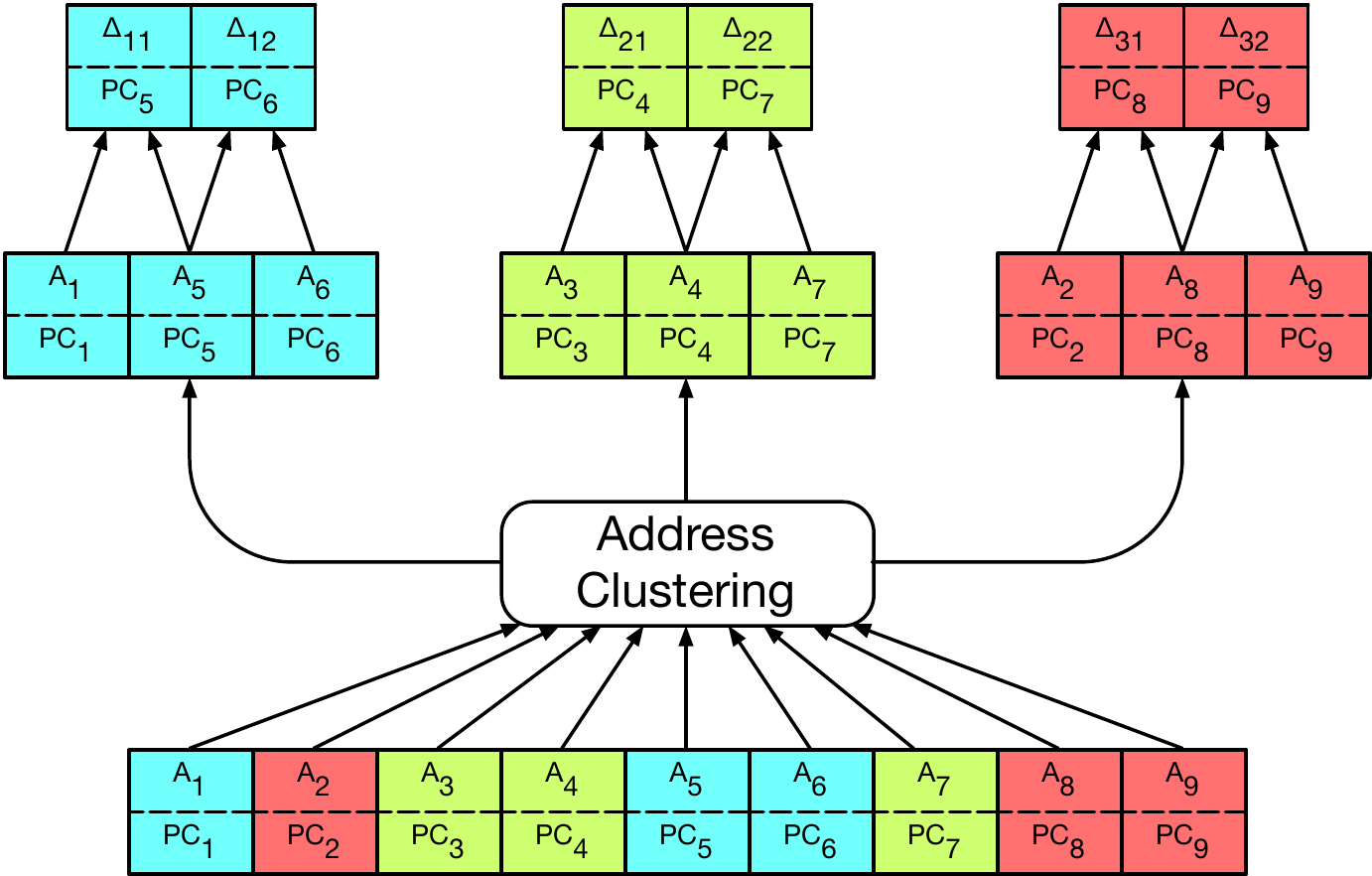}
    	\caption{Clustering the address space into separate streams.}
    	\label{fig:clustering_data_processing}
	\end{subfigure}\begin{subfigure}{0.5\textwidth}
    	\includegraphics[width=\textwidth]{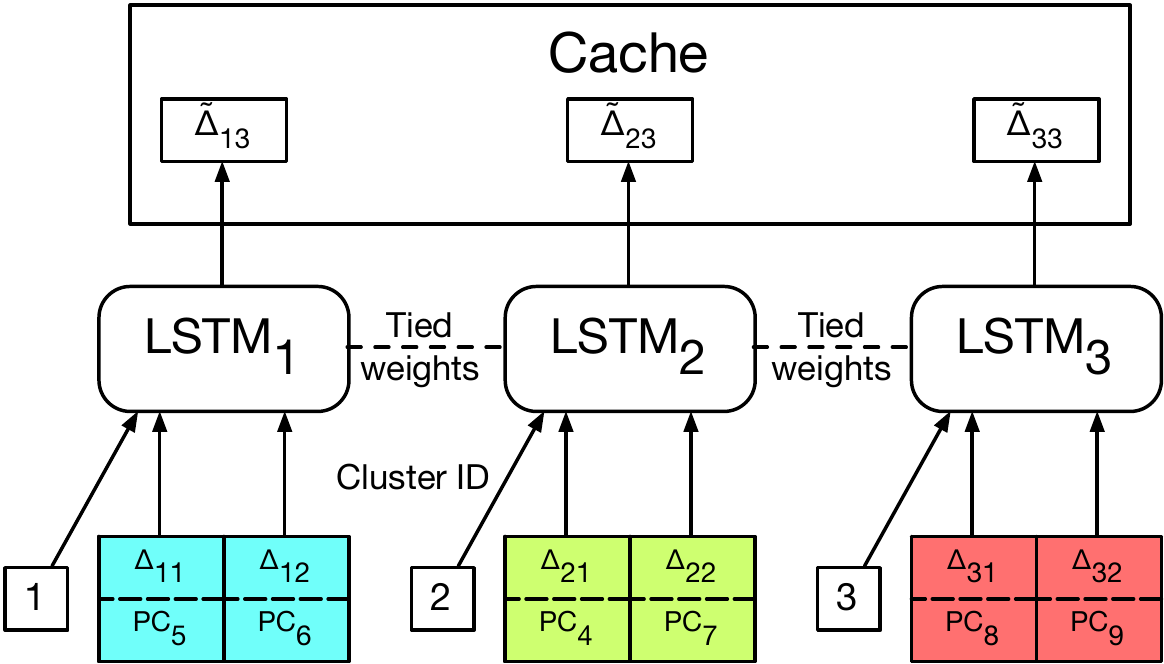}
    	\caption{The clustering + LSTM model.}
	\end{subfigure}
	\caption{The clustering + LSTM data processing and model.}
	\label{fig:clustering_lstm}
\end{figure*}

To assess the relative accuracy of modeling local address-space regions, we first cluster the raw address space using k-means. The data is then partitioned into these clusters, and deltas are computed \textbf{within} each cluster. A visual example of this is shown in Figure~\ref{fig:clustering_data_processing}. We found that one of the major advantages of this approach is that the set of deltas within a cluster is significantly smaller than the global vocabulary, alleviating some of the issues with the embedding LSTM.

To reduce the size of the model, we use a multi-task LSTM to model all of the clusters. Stated another way, we use an LSTM to model each cluster independently, but tie the weights of the LSTMs. However, we provide the \emph{cluster ID} as an additional feature, which effectively gives each LSTM a different set of biases.

The partitioning of the address space into narrower regions also means that the set of addresses within each cluster will take on roughly the same order of magnitude, meaning that the resulting deltas can be effectively normalized and used as real-valued inputs to the LSTM. This allows us to further reduce the size of the model, as we do not need to keep around a large matrix of embeddings. Importantly, we still treat next-delta prediction as a classification problem, as we found that regression is still too inaccurate to be practical~\footnote{The reason for this is after \emph{de-normalization}, small inaccuracies become dramatically magnified.}.

This version of the LSTM addresses some of the issues of the embedding LSTM. The trade-offs are that it requires an additional step of pre-processing to cluster the address space, and that it only models local context. That is, it cannot model the dynamics that cause the program to access different regions of the address space.

\section{Experiments}

A necessary condition for neural networks to be effective prefetchers is that they must be able to accurately predict cache misses. Our experiments primarily measure their effectiveness in this task when compared with traditional hardware. There are many design choices to be made in prefetching beyond the model itself. Creating a fair comparison is a subtle process, and we outline our choices here.

\subsection{Data Collection}

The data used in our evaluation is a dynamic trace that contains the sequence of memory addresses that an application computes. This trace is captured by using a dynamic instrumentation tool, Pin ~\cite{Luk:2005}, that attaches to the process and emits a "PC, Virtual Address" tuple into a file every time the instrumented application accesses memory (every load or store instruction).

This raw access trace mostly contains accesses that hit in the cache (such as stack accesses, which are present in the data cache). Since we are focused on predicting cache misses, we obtain the sequence of cache misses by simulating this trace through a simple cache simulator that emulates an Intel Broadwell microprocessor (Section \ref{sec:pref_prediction}).

To evaluate our proposals, we use the memory intensive applications of SPEC CPU2006. This is a standard benchmark suite that is used pervasively to evaluate the performance of computer systems. However, SPEC CPU2006 also has small working sets when compared to modern datacenter workloads. Therefore in addition to SPEC benchmarks, we also include Google's web search workload. Web search is a unique application that exemplifies enterprise-scale software development and drives industrial hardware platforms.

\subsection{Experimental Setup}

We split each trace into a training and testing set, using 70\% for training and 30\% for evaluation, and train each LSTM on each dataset independently. The embedding LSTM was trained with ADAM~\cite{kingma2014adam} while the clustering LSTM was trained with Adagrad~\cite{duchi2011adaptive}. We report the specific hyperparameters used in the appendix.

\subsection{Metrics}

\paragraph{Precision} We measure precision-at-10, which makes the assumption that each model is allowed to make 10 predictions at a time. The model predictions are deemed correct if the true delta is within the set of deltas given by the top-10 predictions. A label that is outside of the output vocabulary of the model is automatically deemed to be a failure.

\paragraph{Recall} We measure recall-at-10. Each time the model makes predictions, we record this set of 10 deltas. At the end, we measure the recall as the cardinality of the set of predicted deltas over the entire set seen at test-time. This measures the ability of the prefetcher to make diverse predictions, but does not give any weight to the relative frequency of the different deltas.

One subtlety involving the clustering + LSTM model is how it is used at test-time. In practice, if an address generates a cache miss, then we identify the region of this miss, feed it as an input to the appropriate LSTM, and retrieve predictions. Therefore, the bandwidth required to make a prediction is nearly identical between the two LSTM variants.

\subsection{Model Comparison}

We compare our LSTM-based prefetchers to two hardware prefetchers. The first is a standard stream prefetcher. We simulate a hardware structure that supports up to 10 simultaneous streams to maintain parity between the ML and traditional predictors. The second is a GHB PC/DC prefetcher \cite{nes:smith04}. This is a correlation prefetcher that uses two tables. The first table stores PCs, these PCs then serve as a pointer into the second table where delta history is recorded. On every access, the GHB prefetcher jumps through the second table in order to prefetch deltas that it has recorded in the past. This prefetcher excels at more complex memory access patterns, but has much lower recall than the stream prefetcher.

\begin{figure}[ht]
    \centering
    \begin{subfigure}{\columnwidth}
        \includegraphics[width=\textwidth]{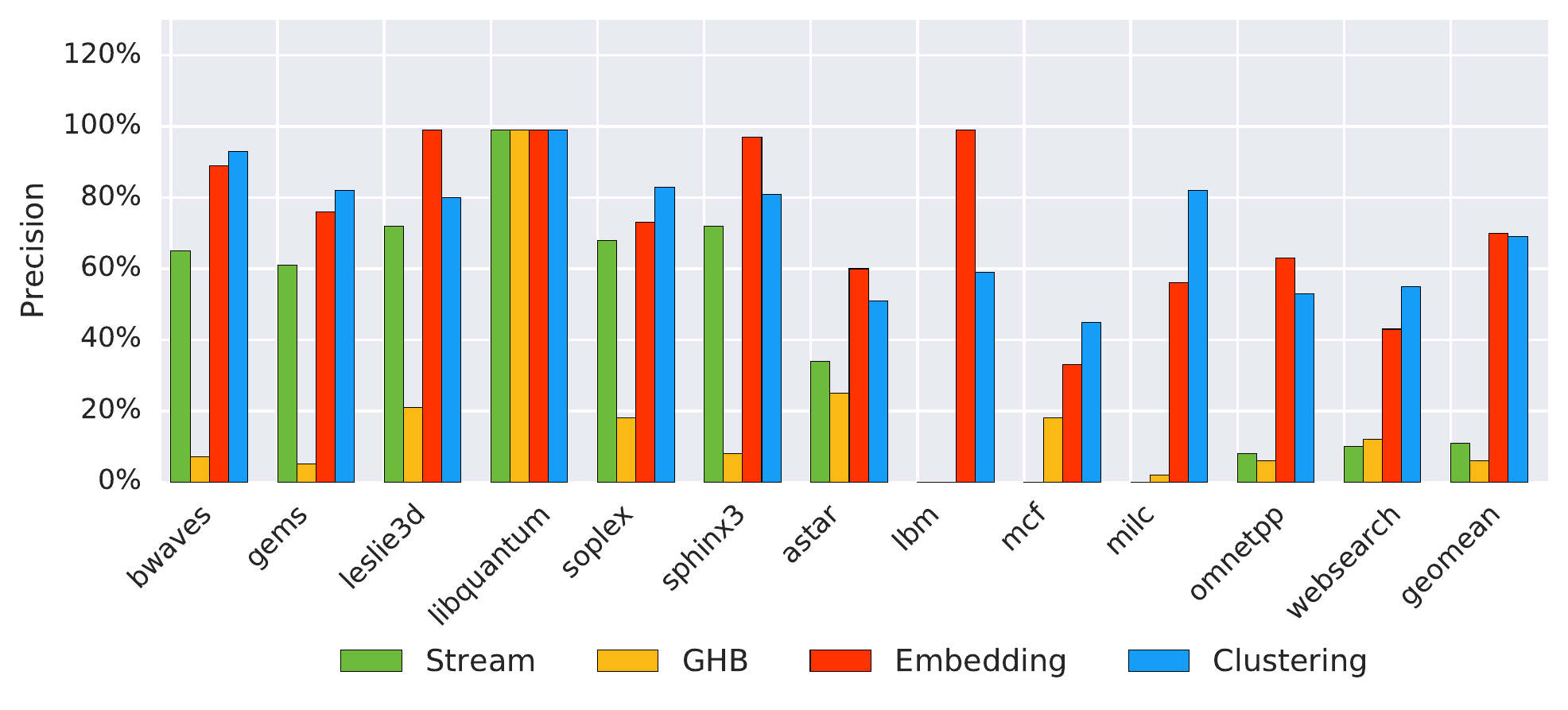}
        \caption{Precision}
    \end{subfigure}
    
    \begin{subfigure}{\columnwidth}
        \includegraphics[width=\textwidth]{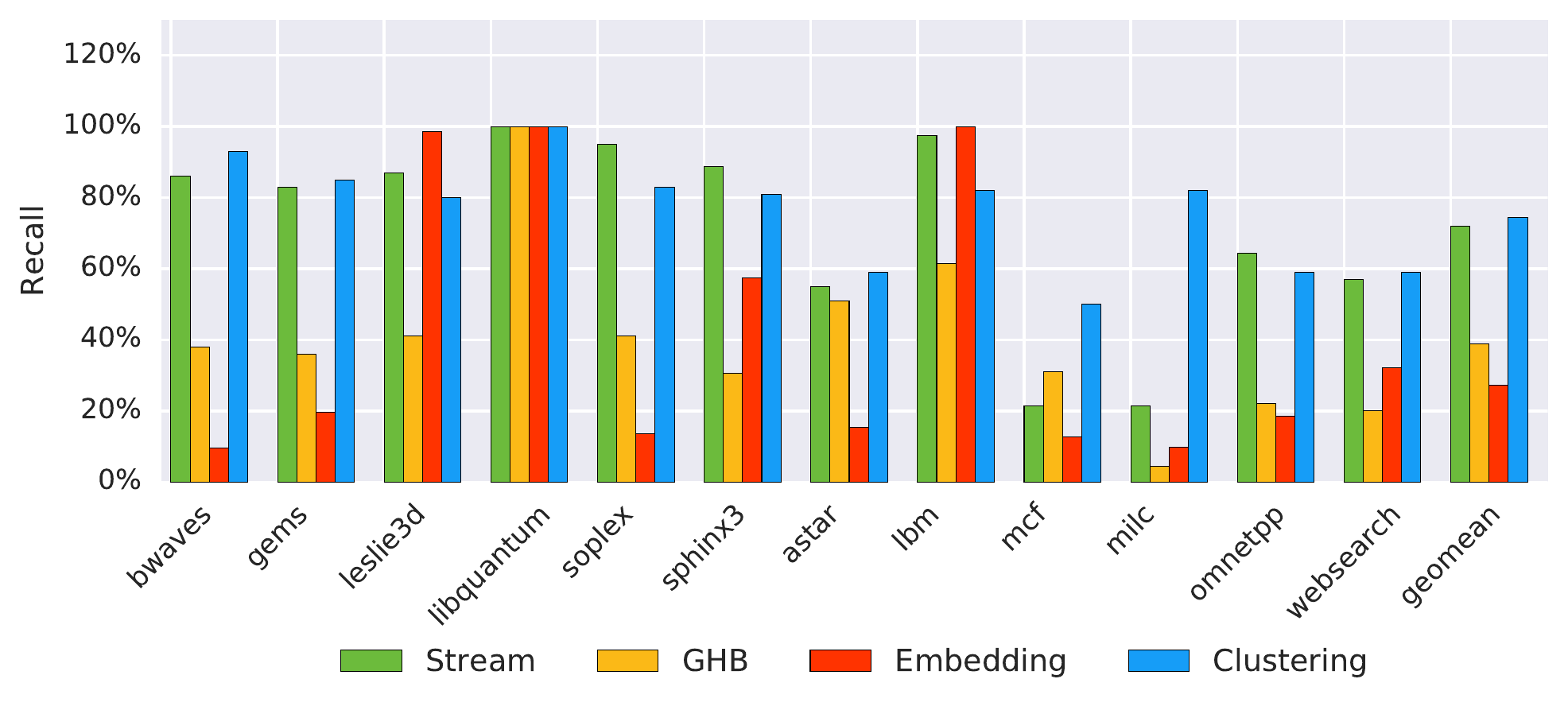}
        \caption{Recall}
    \end{subfigure}
    \caption{Precision and recall comparison between traditional and LSTM prefetchers. Geomean is the geometric mean.}
    \label{fig:model_comparison}
\end{figure}

Figure \ref{fig:model_comparison} shows the comparison of the different prefetchers across a range of benchmark datasets. While the stream prefetcher is able to achieve a high recall due to its dynamic vocabulary, the LSTM models otherwise dominate, especially in terms of precision.

Comparing the embedding LSTM to the cluster + LSTM models, neither model obviously outperforms the other in terms of precision. The clustering + LSTM tends to generate much higher recall, likely the result of having multiple vocabularies. An obvious direction is to ensemble these models, which we leave for future work.

\subsection{Predictive information of $\Delta$s vs PCs}

In this experiment, we remove one of the $\Delta$s or PCs from the embedding LSTM inputs, and measure the change in predictive ability. This allows us to determine the relative information content contained in each input modality.

\begin{figure}[ht]
    \centering
    \begin{subfigure}{\columnwidth}
        \includegraphics[width=\textwidth]{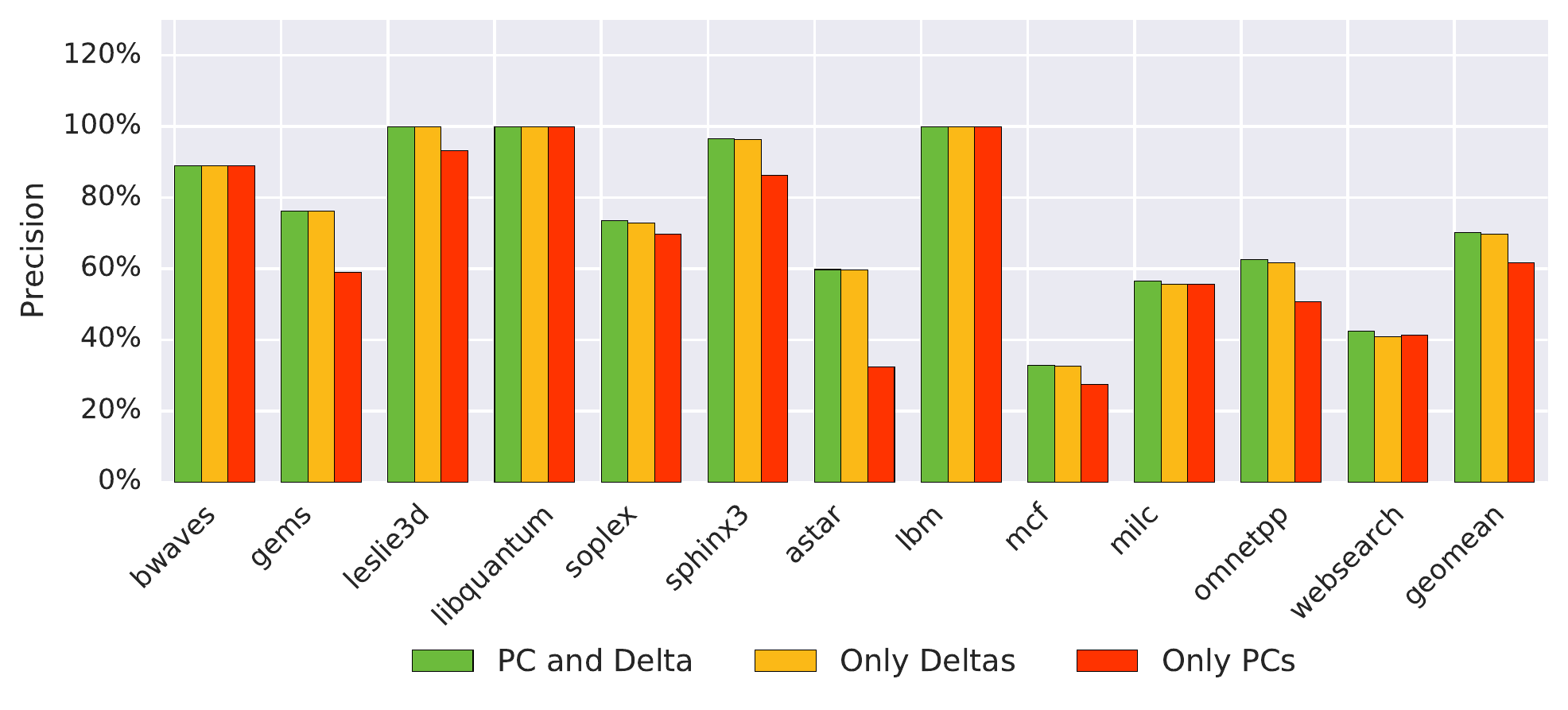}
        \caption{Precision}
    \end{subfigure}
    
    \begin{subfigure}{\columnwidth}
        \includegraphics[width=\textwidth]{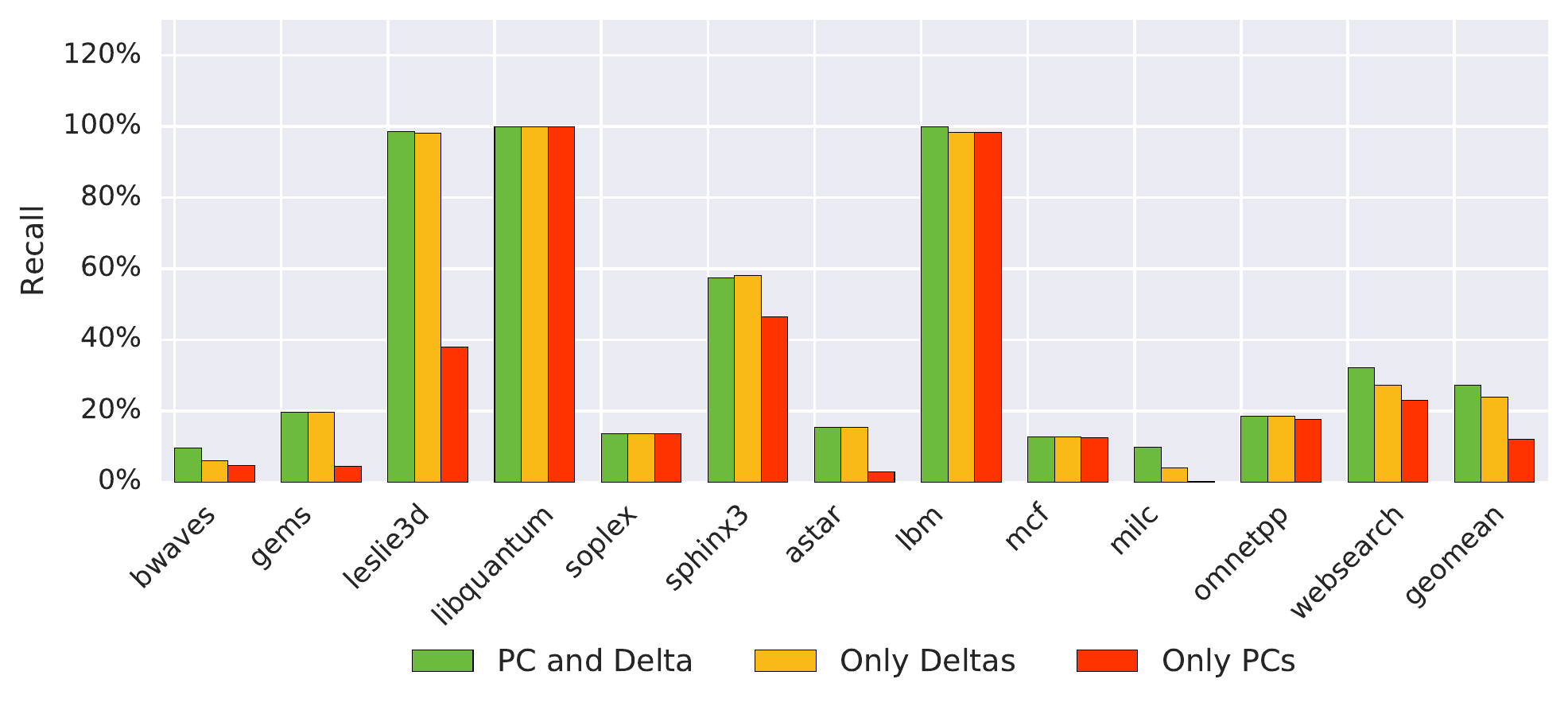}
        \caption{Recall}
    \end{subfigure}
    \caption{Precision and Recall of the embedding LSTM with different input modalities.}
    \label{fig:lstm_ablation}
\end{figure}

As Figure \ref{fig:lstm_ablation} shows, both PCs and deltas contain a good amount of predictive information. Most of the information required for high precision is contained within the delta sequence, however the PC sequence helps improve recall.

\subsection{Interpreting Program Semantics}

\begin{figure}[ht]
    \centering
    \includegraphics[width=\columnwidth]{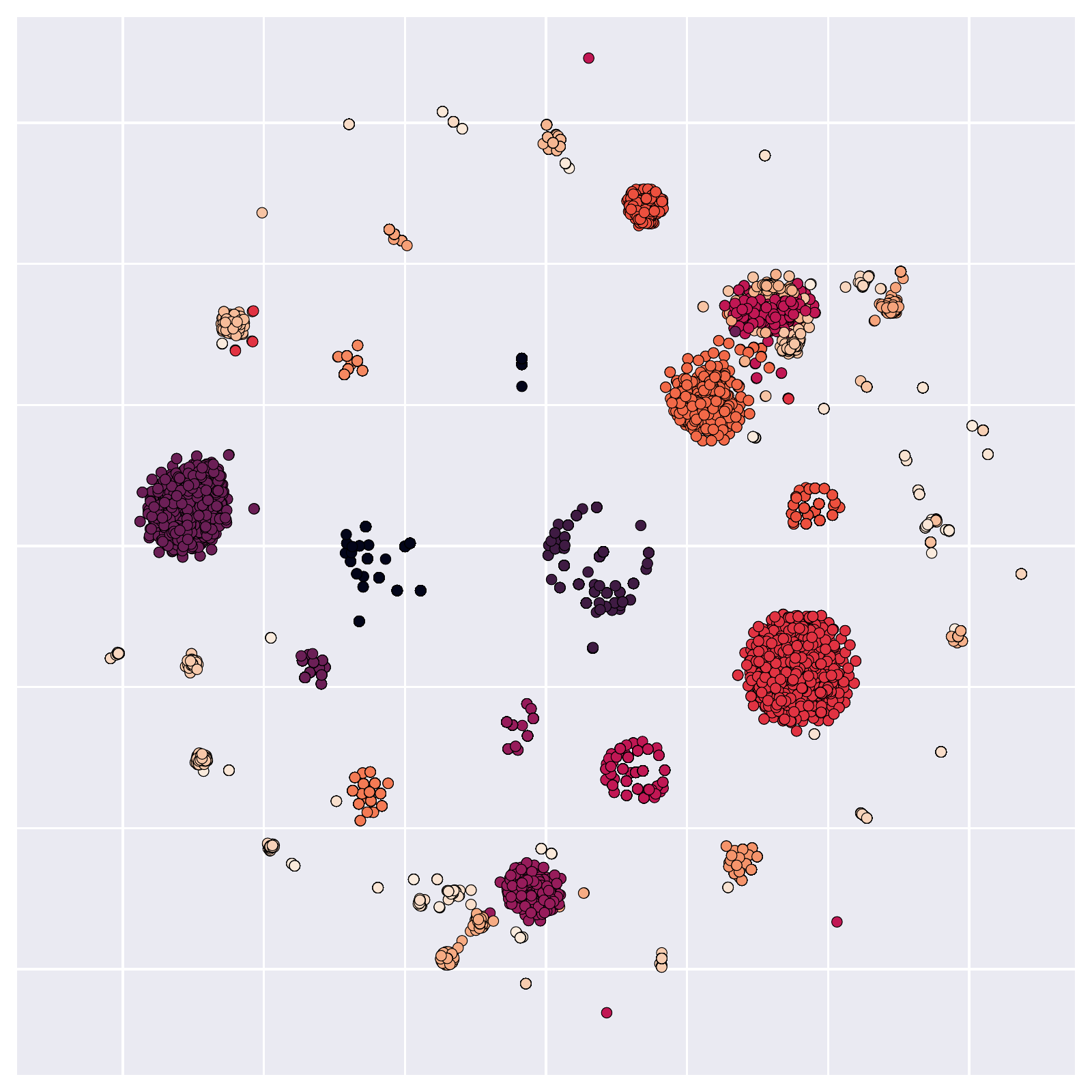}
    \caption{A t-SNE visualization of the concatenated ($\Delta$,PC) embeddings on \textit{mcf} colored according to PC instruction.}
    \label{fig:tsne}
\end{figure}

One of the key advantages of using a model to learn patterns that generalize (as opposed to lookup tables) is that the model can then be introspected in order to gain insights into the data. In Figure \ref{fig:tsne}, we show a t-SNE~\cite{maaten2008visualizing} visualization of the final state of the concatenated ($\Delta$, PC) embeddings on \textit{mcf}, colored according to PCs.

There is clearly a lot of structure to the space. Linking PCs back to the source code in \textit{mcf}, we observe one cluster that consists of repetitions of the same code statement, caused by the compiler unrolling a loop. A different cluster consists only of pointer dereferences, as the application traverses a linked list. Applications besides \textit{mcf} show this learned structure as well. In \textit{omnetpp} we find that inserting and removing into a data structure are mapped to the same cluster and data comparisons are mapped into a different cluster. We show these code examples in the appendix, and leave further inspection for future work, but the model appears to be learning about the higher level structure of the application.

\section{Related Work}

\subsection{Machine Learning in Microarchitecture}
Machine learning in microarchitecture and computer systems is not new, however the application of machine learning as a complete replacement for traditional systems, especially using deep learning, is a relatively new and largely uncharted area. Here we outline several threads of interaction between machine learning and microarchitecture research.

Prior work has also directly applied machine learning techniques to microarchitectural problems. Notably, the perceptron branch predictor \cite{jim:lin01} uses a linear classifier to predict whether a branch is taken or not-taken. The perceptron learns in an online fashion by incrementing or decrementing weights based on taken/not-taken outcome. The key benefit of the perceptron is its simplicity, eschewing more complicated training algorithms such as back-propagation to meet tight latency requirements.

Other applications of machine learning in microarchitecture include applying reinforcement learning for optimizing the long-term performance of memory controller scheduling algorithms \cite{ipek2008self}, tuning performance knobs ~\cite{Blanton:2015}, and using bandits to identify patterns in hardware and software features that relate to a memory access \cite{peled2015semantic}.

Recent work also proposed an LSTM-based prefetcher and evaluated it with generated traces following regular expressions~\cite{zeng2017}. Using the squared loss, the model caters to capturing regular, albeit non-stride, patterns. Irregular memory reference patterns, either due to workload behavior or multi-core processor reordering/interleaving, pose challenges to such regression-based approaches. \citet{zeng2017} evaluate their model on randomly generated patterns. As detailed in Section \ref{sec:problem_formulation}, we have found regression models to be a poor fit on real workloads.

Very recent work has also explored the usage of machine learning to replace conventional database index structures such as b-trees and bloom filters~\cite{kraska2017case}. Although the nature of this problem differs from cache prefetching, there are many similarities as well. Specifically, the idea of using the distribution of the data to learn specific models as opposed to deploying generic data structures. Our clustering approach is also reminiscent of the hierarchical approach that they deploy. Importantly, they find that neural network models are faster to query than conventional data structures.

\subsection{Machine Learning of Program Behavior}

Memory traces can be thought of as a representation of program behavior. Specifically, they represent a bottom-up view of the dynamic interaction of a pre-specified program with a particular set of data. From a machine learning perspective, researchers have taken a top-down approach to explore whether neural networks can understand program behavior and structure.

One active area of research is program synthesis, where a full program is generated from a partial specification--usually input/output examples. \citet{zaremba2014learning} use an LSTM to estimate the output of a randomly generated program.
The model can only see the sequence of ASCII characters representing the program, and must also generate the resulting output as a sequence of characters.
\citet{zaremba2014learning} falls into the category of sequence-to-sequence models~\cite{sutskever2014sequence}. Another example in this category is the \textit{Neural Turing Machine}, which augments an LSTM with an external memory and an attention mechanism to form a differentiable analog of a Turing machine~\cite{graves2014neural}. This is used to solve simple problems such as sorting, copying, and associative recall.

There is also work on modeling source code directly, as opposed to modeling properties of the resulting program. For example, \cite{maddison2014structured} creates a generative model of source code using a probabilistic context-free grammar. \cite{hindle2012naturalness} models source code as if it were natural language using an n-gram model. This approach has been extended to neural language models for source code~\cite{white2015toward}. Lastly, Cummins et al. use neural networks to mine online code repositories to automatically synthesize applications \cite{cummins2017synthesizing}.

\section{Conclusion and Future Work}
\label{sec:conclusion}

Computer architects have long exploited the benefits of learning and predicting program behaviors to unlock control and data parallelism. The conventional approach of table-based predictors, however, is too costly to scale for data-intensive irregular workloads and showing diminishing returns. The models described in this paper demonstrate significantly higher precision and recall than table-based approaches. This study also motivates a rich set of questions that this initial exploration does not solve, and we leave these for future research.

We have focused on a train-offline test-online model, using precision and recall as evaluation metrics. A prefetcher, through accurate prefetching, can change the distribution of cache misses, which could make a static RNN model less effective. There are several ways to alleviate this, such as adapting the RNN online, however this could increase the computational and memory burden of the prefetcher. One could also try training on hits and misses, however this can significantly change the distribution and size of the dataset.

There is a notion of timeliness that is also an important consideration. If the RNN prefetches a line too early, it risks evicting data from the cache that the processor hasn't used yet. If it prefetches too late, the performance impact of the request is minimal, as much of the latency cost of accessing main memory has already been paid. One simple heuristic is to predict several steps ahead, instead of just the next step. This would be similar to the behavior of stream prefetchers.

Finally, the effectivenss of an RNN prefetcher must eventually be measured in terms of its performance impact within a program. Ideally the RNN would be directly optimized for this. This and the previous issues motivate the use of reinforcement learning techniques \cite{sutton1998introduction} as a method to train these RNNs in dynamic environments. Indeed, modern microarchitectures also employ control systems to control prefetcher aggressiveness, and this provides yet another area in which neural networks could be used.

Additionally, we have not evaluated the hardware design aspect of our models. Correlation based prefetchers are difficult to implement in hardware because of their memory size. While it is unclear if RNNs can meet the latency demands required for a hardware accelerator, neural networks also significantly compress learned representations during training, and shift the problem to a compute problem rather than a memory capacity problem. Given the recent proliferation of ML accelerators, this shift towards compute leaves us optimistic at the prospects of neural networks in this domain.

Prefetching is not the only domain where computer systems employ speculative execution. Branch prediction is the process of predicting the direction of branches that an application will take. Branch target buffers predict the address that a branch will redirect control flow to. Cache replacement algorithms predict the best line to evict from a cache when a replacement decision needs to be made. One consequence of replacing microarchitectural heuristics with learned systems is that we can introspect those systems in order to better understand their behavior. Our t-SNE experiments only scratch the surface and show an opportunity to leverage much of the recent work in understanding RNN systems \cite{murdoch2017automatic, james2018beyond}.  

The t-SNE results also indicate that an interesting view of memory access traces is that they are a reflection of program behavior. A trace representation is necessarily different from e.g., input-output pairs of functions, as in particular, traces are a representation of an entire, complex, human-written program. This view of learning dynamic behavior provides a different path towards building neural systems that learn and replicate program behavior.

\bibliography{ml_prefetch}
\bibliographystyle{icml2018}

\clearpage
\appendix
\appendixpage
\lstdefinestyle{mystyle}{
    breakatwhitespace=false,         
    breaklines=true,                 
    captionpos=b,                    
    keepspaces=true,                 
    numbers=left,                    
    numbersep=5pt,                  
    showspaces=false,                
    showstringspaces=false,
    showtabs=false,                  
    tabsize=1,
    frame=single
}
\lstset{style=mystyle}

\section{Interpreting t-SNE Plots}

By mapping PCs back to source code, we observe that the model has learned about program structure. We show examples from two of the most challenging \textit{SPEC CPU2006} applications to learn, \textit{mcf} and \textit{omnetpp}.

\subsection{mcf}

The following function from \textit{mcf} appears in two different t-SNE clusters:

\begin{minipage}{\columnwidth}
    \begin{lstlisting}[language=c++,basicstyle=\ttfamily\small]
    while( node )
    {
        if( node->orientation == UP )
            node->potential = node->basic_arc->cost + node->pred->potential;
        else /* == DOWN */
        {
            node->potential = node->pred->potential - node->basic_arc->cost;
            checksum++;
        }
        tmp = node;
        node = node->child;
        node = tmp;
        
        while( node->pred )
        {
            tmp = node->sibling;
            if( tmp )
            {
                node = tmp;
                break;
            }
            else
                node = node->pred;
        }
    }
    \end{lstlisting}
\end{minipage}

One cluster contains only different instances of line 4, unrolled into three different instructions at three different PCs. We show the line of code, followed by the assembly code in (PC: Instruction) format:

\begin{minipage}{\columnwidth}
    \begin{lstlisting}[language=c++,basicstyle=\ttfamily\small, numbers=none]
    node->potential = node->basic_arc->cost + node->pred->potential;
    401932:	mov  0x18(%rdx),%rsi
    401888:	mov  0x18(%r10),%rsi
    4018df:	mov  0x18(%r11),%rsi
    \end{lstlisting}
\end{minipage}

A second cluster identifies only the PCs responsible for the  linked list traversal, at lines 11 and 16:

\begin{minipage}{\columnwidth}
    \begin{lstlisting}[language=c++, basicstyle=\ttfamily\small, numbers=none]
    node = node->child;
    401878: mov  0x10(%rdx),%r10
    40187c: mov  %rcx,(%rdx)
    
    tmp = node->sibling;
    4019a2: mov 0x20(%r9),%rcx
    \end{lstlisting}
\end{minipage}

\subsection{omnetpp}

\begin{figure}[ht]
    \centering
    \includegraphics[width=0.8\columnwidth]{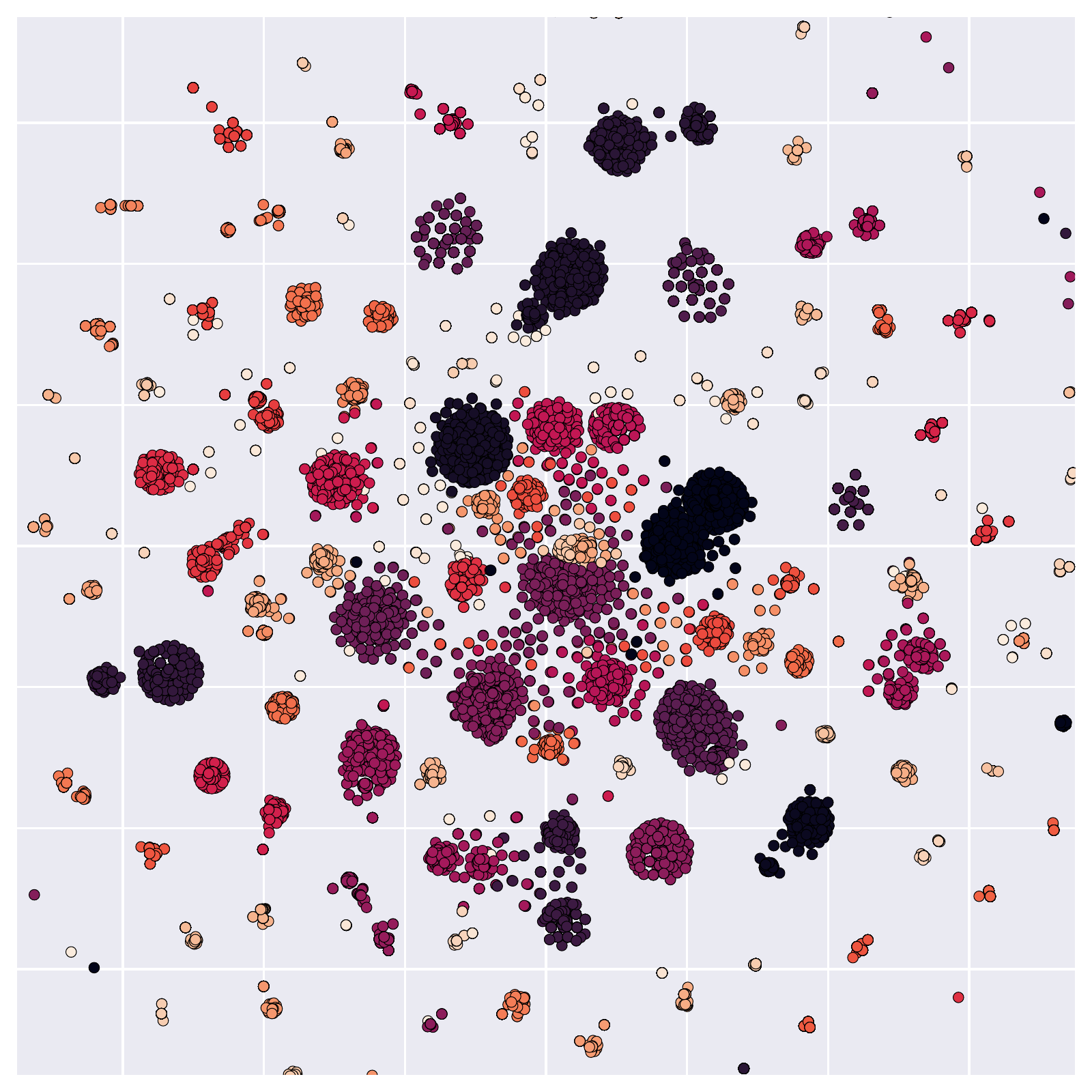}
    \caption{A t-SNE visualization of the concatenated ($\Delta$,PC) embeddings on \ \textit{omnetpp} colored according to PC instruction.}
    \label{fig:tsne_omnet}
\end{figure}

We show the result of running t-SNE on the learned ($\Delta$, PC) embeddings of \textit{omnetpp} in Figure \ref{fig:tsne_omnet}.

\clearpage
Examining some of the clusters closely, we find interesting patterns.
The following code inserts and removes items into an owner's list:

\begin{minipage}{\columnwidth}
    \begin{lstlisting}[language=c++, basicstyle=\ttfamily\small]
    // remove from owner's child list
    if (ownerp!=NULL)   
    {
         if (nextp!=NULL)
              nextp->prevp = prevp;
         if (prevp!=NULL)
              prevp->nextp = nextp;
         if (ownerp->firstchildp==this)
              ownerp->firstchildp = nextp;
         ownerp = NULL;
    }
    // insert into owner's child list as first elem
    if (newowner!=NULL) 
    {
         ownerp = newowner;
         prevp = NULL;
         nextp = ownerp->firstchildp;
         if (nextp!=NULL)
              nextp->prevp = this;
         ownerp->firstchildp = this;
    }
    \end{lstlisting}
\end{minipage}

The main insertion and removal path are both shown in the same t-SNE cluster:

\begin{minipage}{\columnwidth}
    \begin{lstlisting}[language=c++, numbers=none, basicstyle=\ttfamily\small]
    // Removal
    nextp->prevp = prevp;
    448a6a: mov  0x20(%rbx),%r12
    448a6e:	mov  %r12,0x20(%r10)
    //Insertion
    nextp = ownerp->firstchildp;
    44c23a:	mov  0x30(%rax),%r13
    44c23e:	mov  %r13,0x28(%r12)
    \end{lstlisting}
\end{minipage}

\textit{omnetpp}'s t-SNE clusters also contain many examples of comparison code from very different source code files that are used as search statements being mapped to the same t-SNE cluster. Since these comparators are long, they get compiled to many different assembly instructions, so we only show the source code below. Lines 3 and 17 are both mapped to the same t-SNE cluster among other similar comparators:

\begin{minipage}{\columnwidth}
    \begin{lstlisting}[language=c++, basicstyle=\ttfamily\small]
    cObject *cArray::get(int m) 
    {
        if (m>=0 && m<=last && vect[m]) 
            return vect[m];
        else
            return NULL; 
    }
    
    void cMessageHeap::shiftup(int from){
        // restores heap structure (in a sub-heap) 
        int i,j; 
        cMessage *temp; 
        
        i=from; 
        while ((j=2*i) <= n)
        {
        if (j<n && (*h[j] > *h[j+1]))   //direction
            j++;
        if (*h[i] > *h[j])  //is change necessary?
        {
             temp=h[j];
             (h[j]=h[i])->heapindex=j;
             (h[i]=temp)->heapindex=i;
             i=j; 
        }
        else
            break;
        }
    }
    \end{lstlisting}
\end{minipage}

\section{Experimental Results}
\begin{table*}[ht!]
	\begin{center}
	\caption{Experimental Results: Precision}
	\label{table:prec_results}
		\begin{tabular}{ |c|r|r|r|r|r|r| } 
			\hline
			Dataset & Stream & GHB & Embedding & Kmeans & Only PCs & Only Deltas \\
			\hline
			bwaves				&	0.65 & 0.07		& 0.89 & 0.93 & 0.89 & 0.89 \\
			gems				&	0.61 & 0.05		& 0.76 & 0.82 & 0.76 & 0.59 \\
			leslie3d		    &	0.72 & 0.21     & 0.99 & 0.80 & 0.99 & 0.93 \\
			libquantum 		 	&	0.99 & 0.99     & 0.99 & 0.99 & 0.99 & 0.99 \\
			soplex		        &	0.68 & 0.18     & 0.73 & 0.83 & 0.73 & 0.70 \\
			sphinx3	            &	0.72 & 0.08     & 0.97 & 0.81 & 0.96 & 0.86 \\
			astar				&	0.34 & 0.25		& 0.60 & 0.51 & 0.60 & 0.32 \\
			lbm 				&	0.0001 & 0.0001	& 0.99 & 0.59 & 0.99 & 0.99 \\
			mcf			        &	0.0001 & 0.18	& 0.33 & 0.45 & 0.33 & 0.28 \\
			milc 			    &	0.0001 & 0.02	& 0.56 & 0.82 & 0.56 & 0.56 \\
			omnetpp			    &	0.08 & 0.06	 &	0.63 & 0.53 & 0.62 & 0.51 \\
			websearch	        &	0.1  & 0.12  & 0.43 & 0.55 & 0.41 & 0.41 \\
			Geometric Mean	    &	0.11 & 0.06 & 0.70 & 0.69 & 0.70 & 0.61 \\
			\hline
		\end{tabular}
	\end{center}
	\vspace{-.2in}
\end{table*}
\begin{table*}[ht!]
	\begin{center}
	\caption{Experimental Results: Recall}
	\label{table:rec_results}
		\begin{tabular}{ |c|r|r|r|r|r|r| } 
			\hline
			Dataset & Stream & GHB & Embedding & Kmeans & Only PCs & Only Deltas \\
			\hline
			bwaves				&	0.86 & 0.38		& 0.10 & 0.93 & 0.05 & 0.06 \\
			gems				&	0.83 & 0.36		& 0.20 & 0.85 & 0.04 & 0.20 \\
			leslie3d		    &	0.87 & 0.41     & 0.99 & 0.80 & 0.38 & 0.98 \\
			libquantum 		 	&	0.99 & 0.99     & 1.00 & 1.00 & 1.00 & 1.00 \\
			soplex		        &	0.95 & 0.41     & 0.14 & 0.83 & 0.14 & 0.14 \\
			sphinx3	            &	0.89 & 0.30     & 0.57 & 0.81 & 0.46 & 0.58 \\
			astar				&	0.55 & 0.51		& 0.15 & 0.59 & 0.03 & 0.15 \\
			lbm 				&	0.98 & 0.61	& 1.00 & 0.82 & 0.98 & 0.98 \\
			mcf			        &	0.21 & 0.31	& 0.13 & 0.50 & 0.12 & 0.13 \\
			milc 			    &	0.21 & 0.05	& 0.10 & 0.82 & 0.001 & 0.04 \\
			omnetpp			    &	0.64 & 0.22	 &	0.19 & 0.59 & 0.18 & 0.19 \\
			websearch	        &	0.57  & 0.20  & 0.32 & 0.59 & 0.23 & 0.27 \\
			Geometric Mean	    &	0.72 & 0.39 & 0.27 & 0.75 & 0.12 & 0.24 \\
			\hline
		\end{tabular}
	\end{center}
	\vspace{-.2in}
\end{table*}

The experimental results for precision/recall are given in Table \ref{table:prec_results}/Table \ref{table:rec_results} respectively.

\section{LSTM Hyperparameters}
The hyperparameters for both LSTM models are given in Table \ref{table:train_param}
\begin{table*}[h]
  \begin{center}
    \caption{Training hyperparameters for each model.}
    \label{table:train_param}
    \begin{tabular}{| l | l | c |}
      \hline
      Embedding & Network Size & 128x2 LSTM \\
      & Learning Rate & .001 \\
      & Number of Train Steps & 500k \\
      & Sequence Length & 64 \\
      & Embedding Size & 128 \\
      \hline
      Clustering & Network Size & 128x2 LSTM \\
      & Learning Rate & .1 \\
      & Number of Train Steps & 250k \\
      & Sequence Length & 64 \\
      & Number of Centroids & 12 \\ 
      \hline
    \end{tabular}
  \end{center}
\end{table*}

\section{K-Means Clustering on an Address Trace}
\begin{figure*}[htb]
    \centering
    \begin{subfigure}{\columnwidth}
        \includegraphics[width=\textwidth]{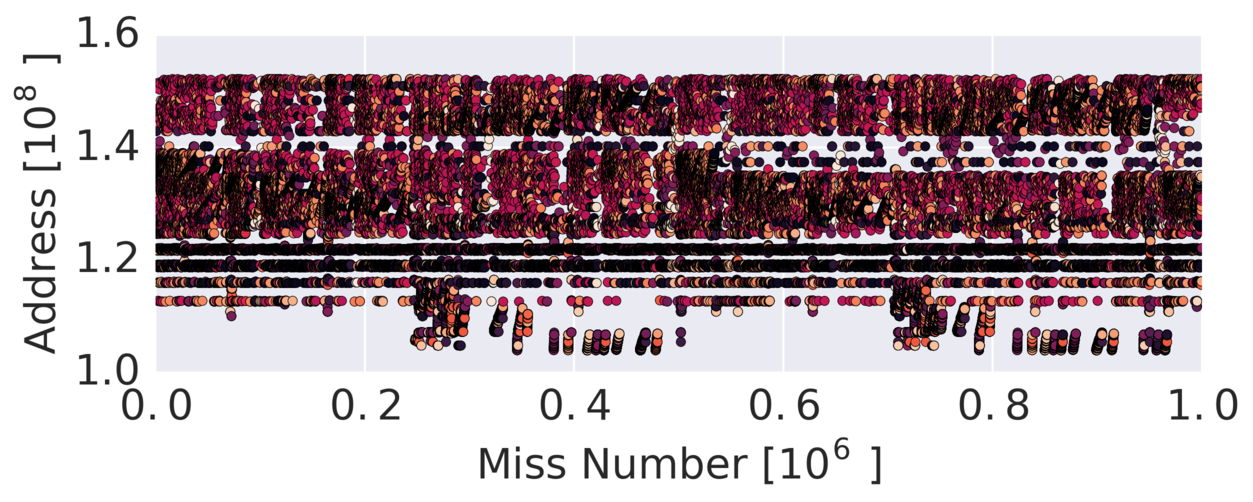}
    \end{subfigure}
    \begin{subfigure}{\columnwidth}
        \includegraphics[width=\textwidth]{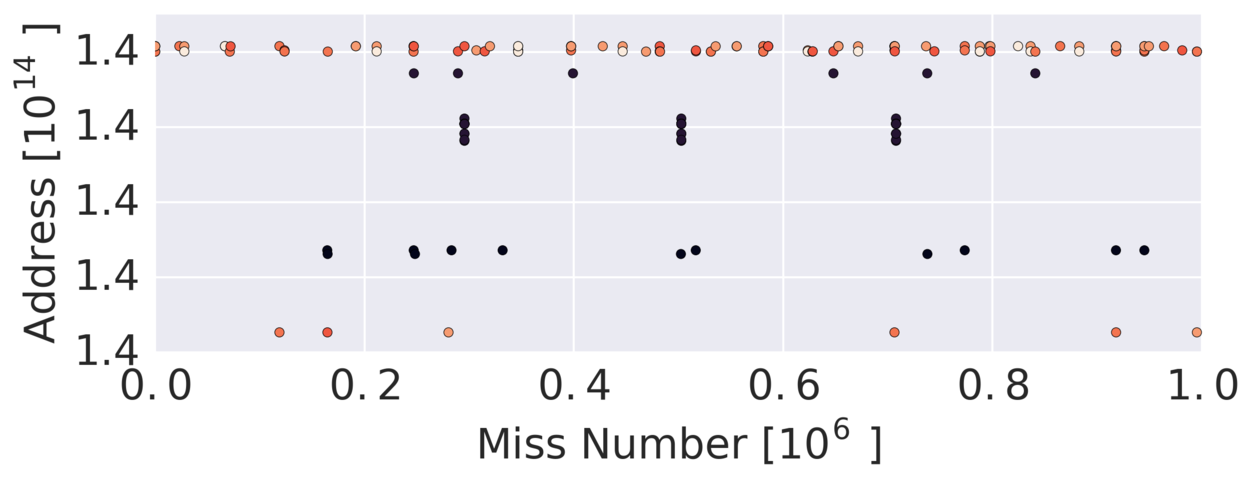}
    \end{subfigure}
    
    \begin{subfigure}{\columnwidth}
        \includegraphics[width=\textwidth]{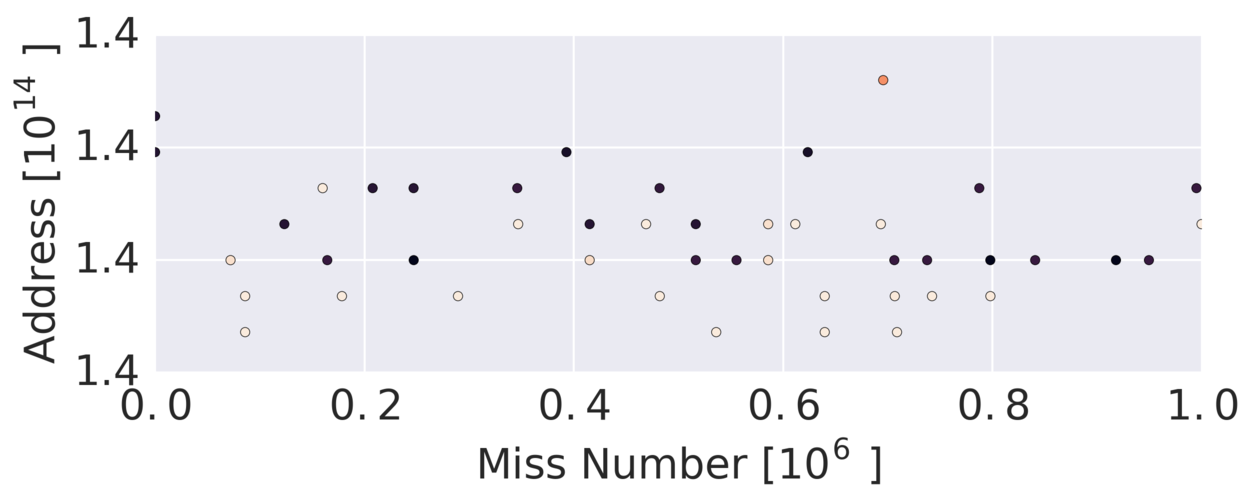}
    \end{subfigure}
    \begin{subfigure}{\columnwidth}
        \includegraphics[width=\textwidth]{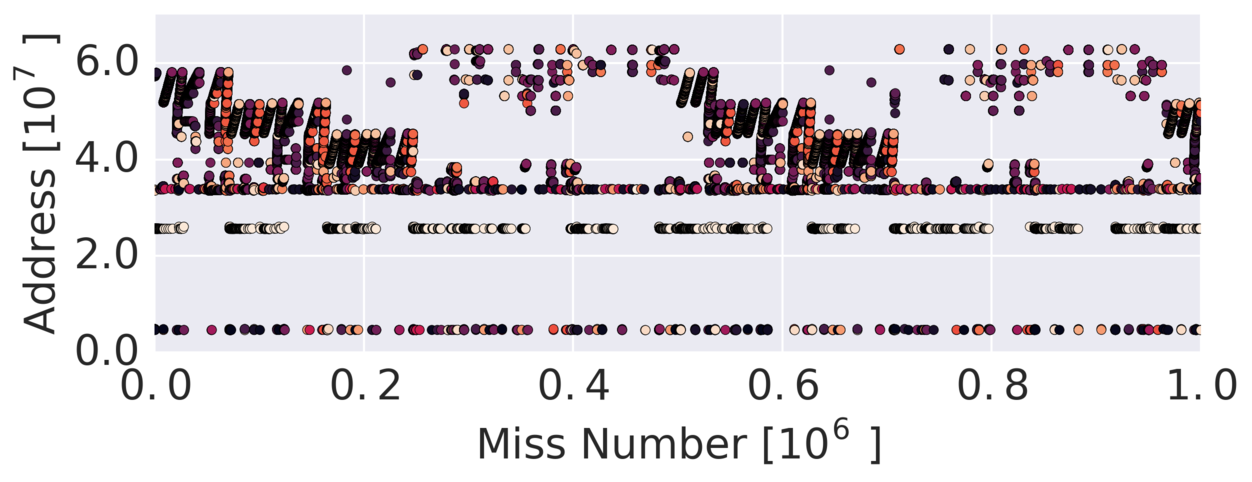}
    \end{subfigure}
    
    \begin{subfigure}{\columnwidth}
        \includegraphics[width=\textwidth]{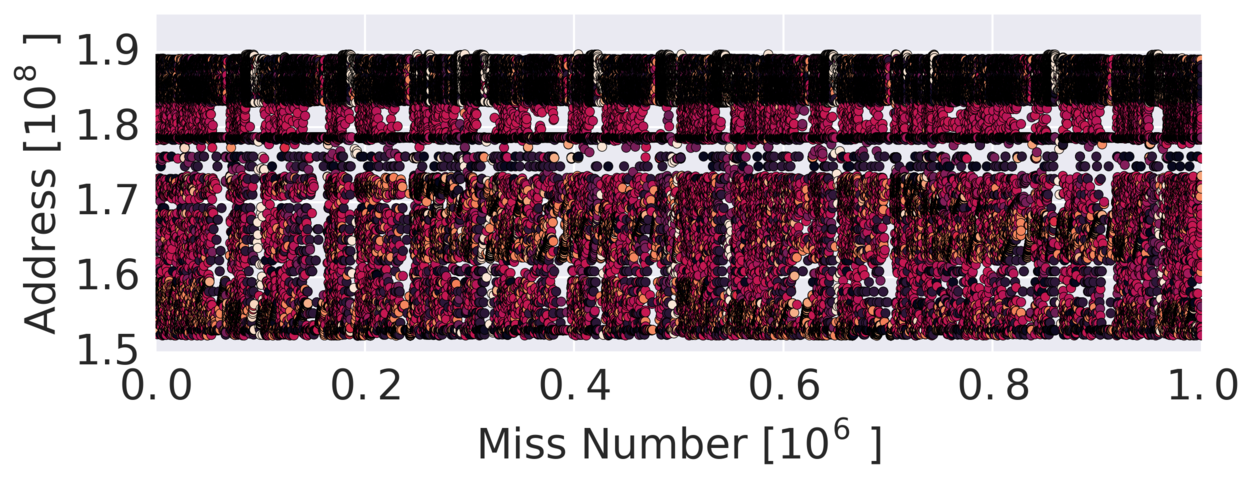}
    \end{subfigure}
    \begin{subfigure}{\columnwidth}
        \includegraphics[width=\textwidth]{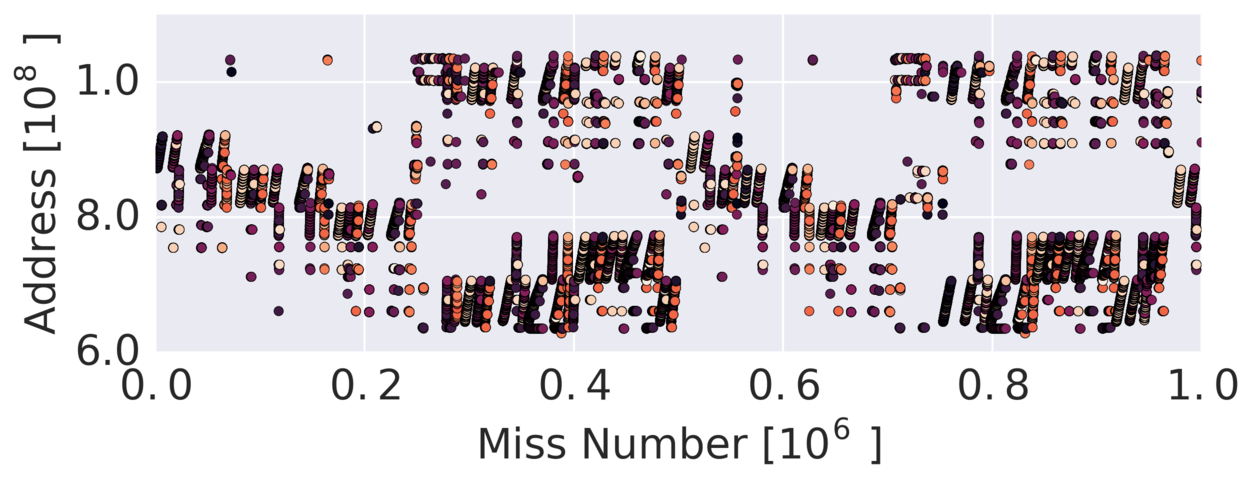}
    \end{subfigure}
    \caption{One million memory accesses from \ \textit{omnetpp} after running k-means clustering on the address space.}
    \label{fig:omnet_clusters_appendix}
\end{figure*}

In Figure \ref{fig:omnet_clusters_appendix} we show the results of running k-means with 6 clusters on $10^6$ addresses from \ \textit{omnetpp}.

\end{document}